\documentclass[10pt,twocolumn,letterpaper]{article}

\usepackage[pagenumbers]{cvpr} % To force page numbers, e.g. for an arXiv version

\usepackage{graphicx}
\usepackage{amsmath}
\usepackage{amssymb}
\usepackage{booktabs}
\usepackage{epsfig}
\usepackage{amsmath}
\usepackage{amssymb}
\usepackage{booktabs}
\usepackage{tabulary}
\usepackage{multirow}
\usepackage{overpic}
\usepackage{xspace}
\usepackage{bbm}
\usepackage{xcolor}         % colors
\usepackage{microtype}
\usepackage{amsmath,amsfonts,bbm}
\usepackage{multirow}
\usepackage{epsfig}
\usepackage{times,amsmath,amssymb,booktabs,tabulary,multirow,overpic,xcolor}
\usepackage{mathtools}
\usepackage{wrapfig}
\usepackage{nicefrac}       % compact symbols for 1/2, etc.
\usepackage{microtype}      % microtypography
\usepackage{enumitem}
\usepackage{bbm}

\usepackage[pagebackref=false, breaklinks=true, letterpaper=true, colorlinks,
            citecolor=citecolor, linkcolor=linkcolor, bookmarks=false]{hyperref}
\definecolor{citecolor}{HTML}{0071BC}
\definecolor{linkcolor}{HTML}{ED1C24}

\definecolor{demphcolor}{RGB}{144, 144, 144}
\newcommand{\demph}[1]{\textcolor{demphcolor}{#1}}

\renewcommand{\paragraph}[1]{\vspace{1.25mm}\noindent\textbf{#1}}

\usepackage[ruled]{algorithm2e}
\usepackage{algorithmic}

\graphicspath{ {./figs/} }
\usepackage{floatrow}
\usepackage{adjustbox}
\newfloatcommand{capbtabbox}{table}[][\FBwidth]

\usepackage[capitalize]{cleveref}
\crefname{section}{Sec.}{Secs.}
\Crefname{section}{Section}{Sections}
\Crefname{table}{Table}{Tables}
\crefname{table}{Tab.}{Tabs.}

\def\paperID{} % *** Enter the CVPR Paper ID here
\def\confName{CVPR}
\def\confYear{2024}

\usepackage{multirow}
\usepackage{cuted}

\newcommand{\vct}[1]{\boldsymbol{#1}} % vector
\newcommand{\mat}[1]{\boldsymbol{#1}} % matrix

\newcommand{\methodname}{{PseCo}\xspace}

\newcommand{\app}{\raise.17ex\hbox{$\scriptstyle\sim$}}

\newlength\savewidth

\usepackage{overpic}

\usepackage[misc]{ifsym}
\newcommand\blfootnote[1]{
    \begingroup
    \renewcommand\thefootnote{}\footnote{#1}
    \addtocounter{footnote}{-1}
    \endgroup
}
\usepackage[hang]{footmisc}
\title{
Point, Segment and Count: A Generalized Framework for Object Counting
}

\author{Zhizhong Huang$^{1}$\qquad Mingliang Dai$^{1}$\qquad Yi Zhang$^{2}$\qquad Junping Zhang$^{1}$\qquad Hongming Shan$^{3\dagger}$
\\
$^{1}$ Shanghai Key Lab of Intelligent Information Processing, School of Computer Science,\\
Fudan University, Shanghai 200433, China\\
$^{2}$ School of Cyber Science and Engineering, Sichuan University, Chengdu 610065, China\\
$^{3}$ Institute of Science and Technology for Brain-inspired Intelligence \& MOE Frontiers Center for \\Brain Science \& Key Laboratory of Computational Neuroscience and Brain-Inspired
Intelligence,\\
Fudan University, Shanghai 200433, China\\
{\tt\small \{zzhuang19, mldai21, jpzhang, hmshan\}@fudan.edu.cn, yzhang@scu.edu.cn} \\
{\small Code:~\url{https://github.com/Hzzone/PseCo}}
}

\begin{document}
\maketitle
\blfootnote{$\dagger$ Corresponding author.}
\thispagestyle{empty}

\begin{abstract}
Class-agnostic object counting aims to count all objects in an image with respect to example boxes or class names, \emph{a.k.a} few-shot and zero-shot counting. 
% Current state-of-the-art methods highly rely on density maps to predict object counts, which lacks model interpretability.
In this paper, we propose a generalized framework for both few-shot and zero-shot object counting based on detection.
Our framework combines the superior advantages of two foundation models without compromising their zero-shot capability: (\textbf{i}) SAM to segment all possible objects as mask proposals, and (\textbf{ii}) CLIP to classify proposals to obtain accurate object counts. However, this strategy meets the obstacles of efficiency overhead and the small crowded objects that cannot be localized and distinguished.
To address these issues, our framework, termed \methodname, follows three steps: point, segment, and count.
Specifically, we first propose a class-agnostic object localization to provide accurate but least point prompts for SAM, which consequently not only reduces computation costs but also avoids missing small objects.
Furthermore, we propose a generalized object classification that leverages CLIP image/text embeddings as the classifier, following a hierarchical knowledge distillation to obtain discriminative classifications among hierarchical mask proposals.
Extensive experimental results on FSC-147, COCO, and LVIS demonstrate that \methodname achieves state-of-the-art performance in both few-shot/zero-shot object counting/detection.
\end{abstract}
\section{Introduction}

Object counting has attracted growing research interest in recent years. It aims to estimate the specific object counts in an image, especially in extremely crowded scenes that cannot be distinguished or counted one-by-one by humans. Traditional object counting methods typically focus on specific categories such as humans~\cite{dai2023cross}, animals, or cars. One well-known direction is crowd counting which counts all presented persons in an image. However, it requires a labor-intensive amount of training data with point annotations and is limited to the pre-defined category once the model is trained. Therefore, the recent efforts in object counting resort to class-agnostic counting, which counts arbitrary categories with a few guidance from example images~\cite{lu2019class,ranjan2021learning,fan2020few,liu2022countr,shi2022represent,nguyen2022few,djukic2023low} and class names~\cite{ranjan2022exemplar,xu2023zero,shi2023training}. It achieves satisfactory performance even though the categories are unseen during training, which thus reduces the burden of data starvation.

\begin{figure}[t]
    \centering
    \includegraphics[width=1.0\linewidth]{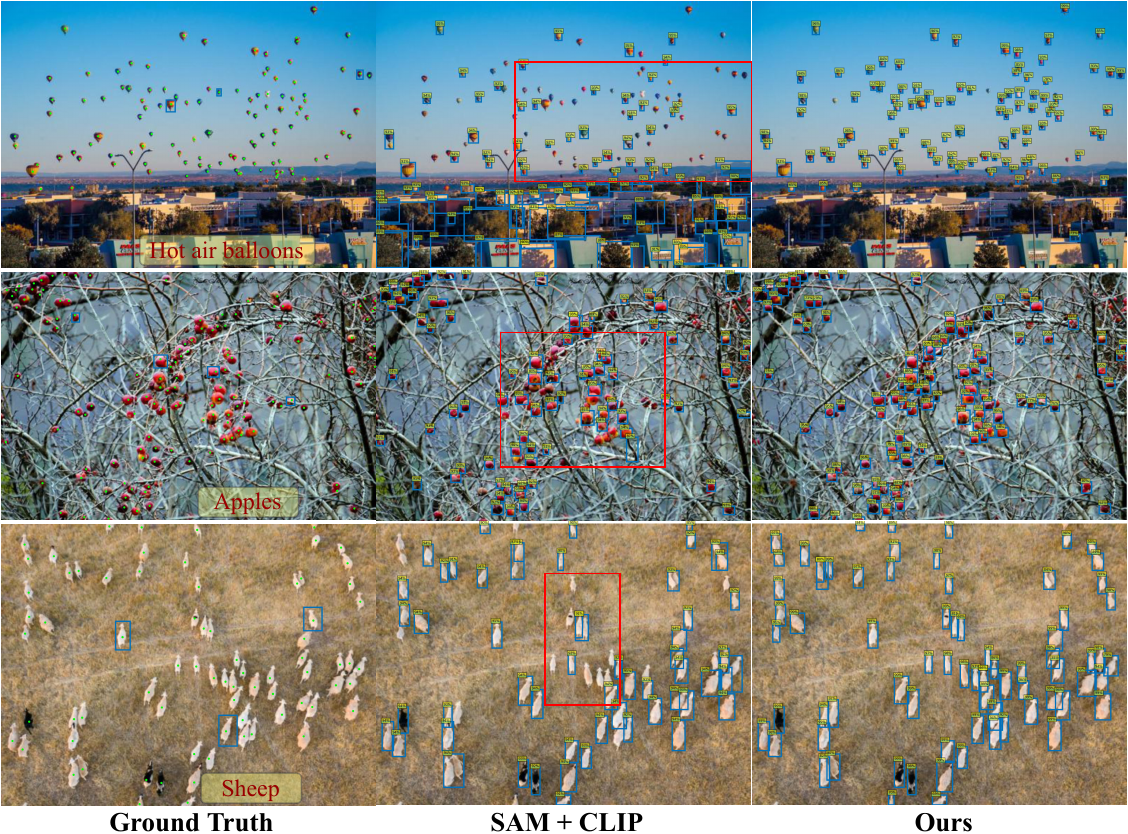}
    \caption{
    Sample results of vanilla SAM + CLIP and the proposed method. Given the class name~(zero-shot) or example boxes~(few-shot), our method can detect all objects in the image for counting.
    \label{fig:motivation}
    }
\end{figure}

Most of class-agnostic object counting methods involve generating density maps,  which can be summed to derive the object counts. Typically, they compute the similarity between visual features of input and example images to guide the object counting~\cite{ranjan2021learning,liu2022countr,shi2022represent,djukic2023low}, \ie, few-shot object counting if examples are provided. In contrast, zero-shot object counting~\cite{xu2023zero} only uses class names to select the best examples in the image. In summary, they mainly focus on how to improve the quality of similarity maps to produce density maps through better examples~\cite{xu2023zero}, transformer~\cite{liu2022countr}, and attention~\cite{djukic2023low}. However, these density-based methods lack interpretability as density maps are hard to verify. On the other hand, detecting target objects for counting is a potential solution but box annotations are much more cumbersome to collect than points. Although the literature explores using only a few boxes and all point annotations~\cite{nguyen2022few}, it presents inferior performance on both tasks as the model usually overfits training categories~\cite{wang2020frustratingly}.

Recent years have witnessed a great breakthrough in the foundation models of computer vision, \eg, Segment Anything Model~(SAM)~\cite{kirillov2023segment} for  segmentation and Contrastive Language-Image Pre-Training (CLIP)~\cite{radford2021learning} for the vision language model. Both of them have shown great zero-shot potential in generalizing to novel scenes. A simple solution for object counting is to employ SAM to segment everything in an image and use CLIP to classify the regions with respect to given examples/texts. However, it remains challenging to combine their advantages for object counting. Fig.~\ref{fig:motivation} reveals the problems of directly applying SAM and CLIP to counting. 
\emph{First}, the small objects tend to be missed by SAM, which cannot be localized by uniform grid point prompts.
\emph{Second}, it is still time-consuming if directly using CLIP to classify the cropped image regions.
\emph{Third}, object counting needs more discriminative classifications, especially for small objects, otherwise most of them cannot be distinguished from the background.

In this paper, we propose a generalized framework for object counting, termed \methodname, to address these issues in the following aspects.
\emph{First}, instead of using a predefined uniform grid point prompts for SAM to segment everything in an image, we propose a class-agnostic object localization that estimates a heatmap of all objects, from which we can infer the object coordinates of each object. Subsequently, it can provide accurate but least point prompts for SAM, segmenting small objects while reducing the computation costs.
\emph{Second}, we propose to leverage the CLIP text/image embeddings to classify the image regions, formulating a generalized framework for both zero-shot and few-shot object counting. Hence, our framework can detect and count arbitrary classes using examples/texts.
\emph{Third}, we propose a hierarchical knowledge distillation to distill the zero-shot capabilities of CLIP to our \methodname. It discriminates among the hierarchical mask proposals produced by SAM, enabling our \methodname to distinguish the desired objects from a large number of proposals.
Fig.~\ref{fig:motivation} presents the results of \methodname, which effectively detects and distinguishes the target objects, including those that are very small.

The contributions are summarized as follows:
\begin{itemize}
[leftmargin=*]
\setlength{\itemsep}{0pt} 
\setlength{\parskip}{0pt}%
\setlength{\topsep}{0pt}%
\item We present \methodname, a generalized framework that leverages the advantages of SAM and CLIP for both few-shot and zero-shot object detection and counting.
\item We propose a novel class-agnostic object localization, which can provide sufficient, accurate but least point prompts for SAM to segment all possible objects.
\item We further propose a generalized object classification to identify target objects, with a hierarchical knowledge distillation to learn discriminative classification.
\item Experimental results on various benchmarks, including FSC-147, COCO, and LVIS datasets, demonstrate the effectiveness of \methodname on object detection/counting.
\end{itemize}
\section{Related Work}
\paragraph{Class-agnostic object counting.}
Traditional object counting focuses on specific categories such as car~\cite{mundhenk2016large} and human~\cite{lian2019density,zhang2016single,gao2023deep,wang2020nwpu,goldman2019precise,hsieh2017drone}, which can be divided into density-based and detection-based methods. Density-based methods~\cite{lian2019density,zhang2016single,gao2023deep} predict and sum over density maps to infer the counting results. The detection-based methods~\cite{goldman2019precise,hsieh2017drone} resort to object detection for counting. The former performs well in crowded scenes, and the latter provides better interpretability but needs box annotations. The class-agnostic counting~\cite{cholakkal2019object,xu2023zero,ranjan2022exemplar,shi2023training,djukic2023low,fan2020few,liu2022countr,shi2022represent,lu2019class,jiang2023clip} is not limited to specific categories. Instead, they count the target objects through some exemplar bounding boxes of a new category~(few-shot)~\cite{gong2022class,shi2023training,djukic2023low,fan2020few,liu2022countr,shi2022represent,lu2019class}, or a class name~(zero-shot)~\cite{xu2023zero,jiang2023clip}. Most of them compute the similarity maps between visual features of images and examples to infer density maps. Although zero shot,~\cite{xu2023zero} selects the best proposals with respect to class names to construct examples. C-DETR~\cite{fan2020few} detects objects only trained on point and a few box annotations. SAM-Free~\cite{shi2023training} generates mask prior from grid points and selects better point prompts from the CLIP similarity map without training.
However, it is inferior, especially for small objects.
% However, it produces inferior results, especially for small objects.

\paragraph{CLIP-based object detection.}
CLIP~\cite{radford2021learning} learns well-aligned text-image embeddings, and can be applied to object detection~\cite{zareian2021open,gu2021open,zhou2022detecting,zang2022open,zhong2022regionclip,zhang2023simple}, segmentation~\cite{9616392,What_Transferred_Dong_CVPR2020,dong2023federated_FISS}, adversarial attack~\cite{yunlong2024attack}, and generation~\cite{wei2023dreamvideo}. Typically, ViLD~\cite{gu2021open} distills the knowledge of CLIP to the region classifier of Mask R-CNN~\cite{he2017mask} so that CLIP can extract the classification weights from text or image for any novel class. RegionCLIP~\cite{zhong2022regionclip} retrains CLIP on the data of region-text pairs. OV-DETR~\cite{zang2022open} is transformed into a binary matching problem conditioned on CLIP embeddings.
CLIP enables these methods with zero-shot capability for open-vocabulary object detection. In contrast, the proposed \methodname is a generalized framework for few-shot/zero-shot object detection and uses a novel hierarchical knowledge distillation to encourage discriminative classifier between mask proposals.

\begin{figure*}[t]
    \centering
    \includegraphics[width=1.00\linewidth]{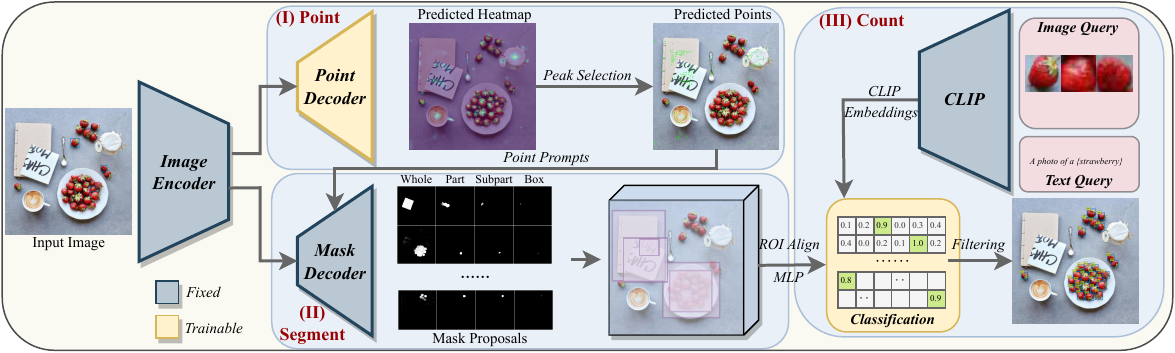}
    \caption{
    Illustration of the proposed \methodname, following the steps: point, segment, and count.
    Given an input image, the point decoder predicts the class-agnostic heatmap to point out all objects. The image encoder and mask decoder from SAM are fixed during training~(the prompt encoder is omitted here) and output the mask proposals. The proposals are classified with respect to CLIP image/text embeddings.
    \label{fig:framework}
    }
\end{figure*}
\section{The Proposed Approach}

\subsection{Preliminaries}
\textbf{SAM}~\cite{kirillov2023segment} consists of three key components: \textbf{(i)} an image encoder to extract image features; \textbf{(ii)} a prompt encoder to encode the user-input prompts; and \textbf{(iii)} a mask decoder to predict the segmentation masks under the control of given prompts. In particular, it utilizes points and boxes as prompts to generate high-quality hierarchical object masks, including whole, part, and subpart segmentations of each prompt. Due to massive training data, SAM exhibits promising zero-shot segmentation performance on various benchmarks. In particular, SAM receives uniform grid point prompts to segment every possible object in an image.

\textbf{CLIP}~\cite{radford2021learning} learns well-aligned vision-language representations with contrastive loss from large-scale image-text pairs. It contains two separate encoders for each modality but maps the data into the same embedding space. In zero-shot classification, CLIP builds the image classifier using a predefined template, \eg, `A photo of dog' when only the class name `dog' is available.

\textbf{Two-stage object detection} such as Faster R-CNN~\cite{ren2015faster} divide the object detection into two stages. The first stage uses a region proposal network~(RPN) to produce a coarse set of object boxes~(proposals) and class-agnostic objectness scores. The second stage takes these proposals for classification and refines the box coordinates. In particular, open-vocabulary object detection~\cite{gu2021open,zhou2022detecting,zhong2022regionclip,zhang2023simple} focuses on improving zero-shot classification of the second stage, which uses CLIP embeddings as the classification weights.

SAM and CLIP have shown their potential in zero-shot segmentation and classification. In this paper, we study a challenging problem about how to synergize their advantages for object counting under a similar framework of two-stage object detection, without compromising their zero-shot capabilities when generalizing to novel scenes.

\subsection{Problem Formulation and Framework}
\label{sec:problem_formulation}
As presented in Fig.~\ref{fig:framework}, given an input image, our goal is to count the target objects with respect to a set of image/text queries. Instead of predicting a density map, we formulate the object counting as object detection; that is, detect and  count them all. 

Inspired by two-stage object detection, we build our framework into the following steps: point, segment, and count as shown in Fig.~\ref{fig:framework} under the help of SAM and CLIP.
Specifically, \methodname (\textbf{i}) points out all possible objects using least point coordinates,
(\textbf{ii}) uses SAM to generate the corresponding mask proposals conditioned on the point prompts, 
and (\textbf{iii}) classifies and post-processes all proposals to count target objects.
In a sense, point and segment steps perform a similar role as the RPN in Faster R-CNN to provide sufficient class-agnostic object proposals for the subsequent object classification. On the other hand, the counting step can filter the undesired proposals by thresholding the scores with respect to examples/texts. At this step, we can further use non-maximum suppression~(NMS) to remove duplicate proposals like object detection.

\paragraph{A simple baseline.}
Following the spirit of point, segment, and count, we design a simple baseline that leverages both SAM and CLIP. Uniform grid points, \eg, $32\times 32$ are used as prompts for SAM to segment all objects. The image regions are cropped from the input image by predicted proposals, which are then fed into CLIP for classification.

However, such a simple baseline has the following limitations. \emph{First}, $32\times 32$ grid points may be insufficient to enumerate all objects, especially under crowded scenes. Consequently, many small objects could be ignored, which is not suitable for object counting. Although increasing the number of points can alleviate this problem, it could inevitably waste heavy computational costs as many points are located in the background, which is impractical in real-world applications. \emph{Second}, CLIP can only produce image-level representations, and it is computationally expensive to crop proposals during inference time. Although the vision encoder of CLIP can be distilled~\cite{gu2021open}, the representations are not discriminative enough for small regions, as shown in Fig.~\ref{fig:motivation}.

To address the above problems, we propose a novel framework in Fig.~\ref{fig:framework}, termed \methodname, for generalized object counting and detection. Specifically, \methodname only trains a point decoder to point out all objects using the least points under the setting of keypoint estimation and a classifier to classify all proposals. Both of them are built upon the image features extracted from the pre-trained image encoder of SAM, leading to negligible computational costs.

We detail their designs in the following.

\subsection{Class-agnostic Object Localization}
\label{sec:object_localization}

\begin{figure}[t]
    \centering
    \includegraphics[width=1.0\linewidth]{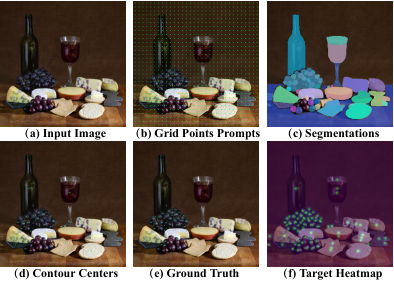}
    \caption{
    Sample results to generate the class-agnostic target heatmaps. Given (a) input image and (b) uniform grid point prompts, SAM predicts all (c) segmentation. We combine (d) all contour centers of segmentations
    to avoid bad point prompts
    and (e) ground-truth point annotations to produce (f) target heatmap. The resultant heatmap will be used to supervise the point decoder.
    \label{fig:heatmap_generation}
    }
    \vspace{-5pt}
\end{figure}

\begin{figure*}[t]
    \centering
    \includegraphics[width=1.0\linewidth]{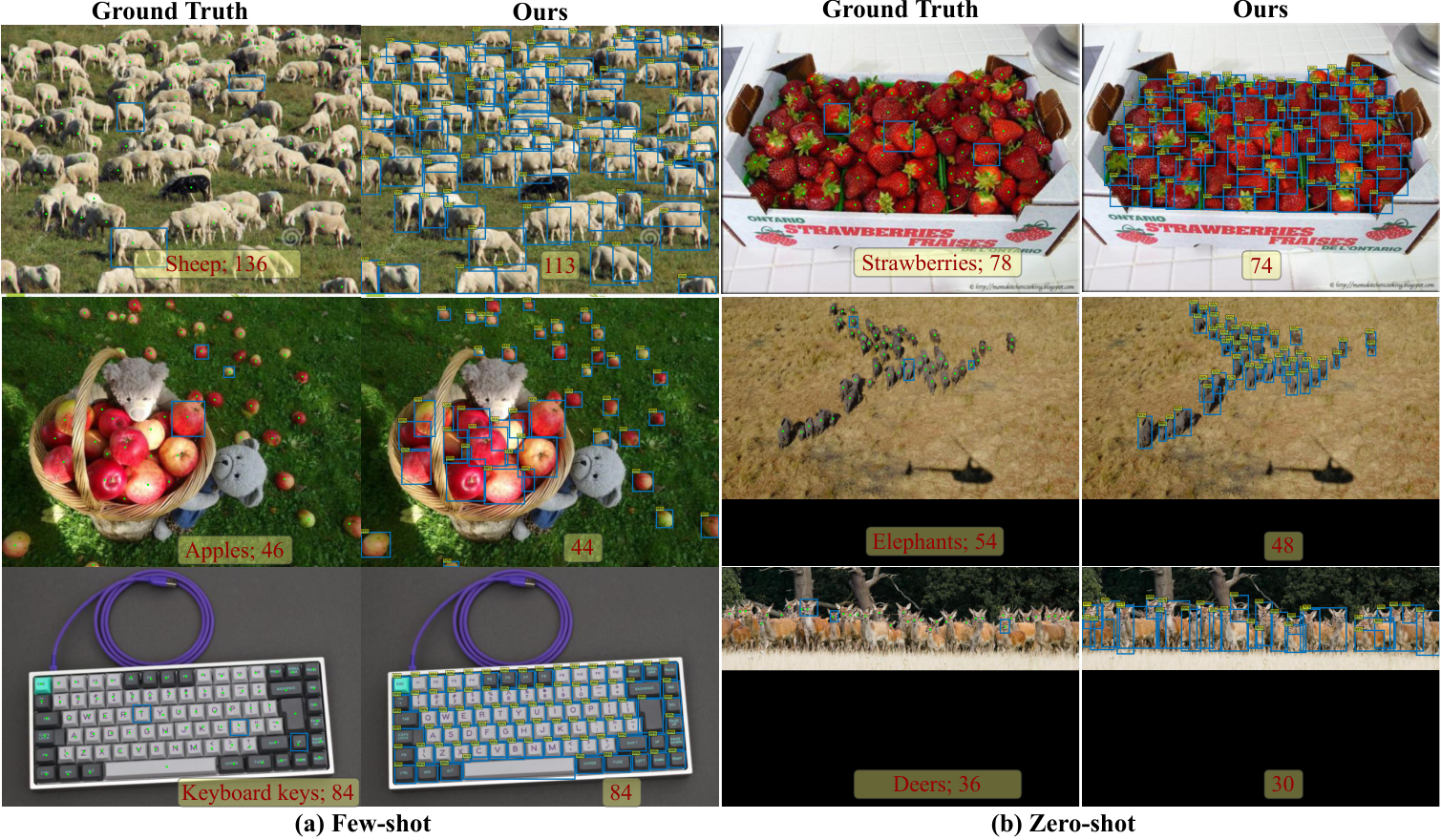}
    \caption{
        Qualitative results for (a) few-shot and (b) zero-shot object counting and detection. The class names, ground-truth counts, and our predicted counts are in color boxes. Zoom in for better view.
    }
    \vspace{-5pt}
    \label{fig:qualitative_results}
\end{figure*}

As presented in Fig.~\ref{fig:heatmap_generation} (a), (b), (c), SAM ignores some small objects using uniform grid point prompts.
To localize all possible objects with the least points, we propose to formulate this problem as keypoint estimation. Let $\mat{I}\in \mathbb{R}^{H\times W}$ be the input image, the point decoder aims to produce the class-agnostic keypoint heatmap $\widehat{\mat{H}}\in [0, 1]^{\frac{H}{s}\times \frac{W}{s}}$, where $s$ is output stride. In SAM, $\mat{I}$ is resized to $H=W=1024$ along the longest side, and the point decoder shares the same architecture as the mask decoder with $s=4$.

Although ground-truth points are available during training, the point decoder is prone to overfit the categories in the training data~\cite{wang2020frustratingly}, which cannot generalize to zero-shot scenarios that the categories during testing are unseen by the model. In other words, the point annotations do not contain all possible objects in the image, and thus the novel categories could be misclassified as background. Benefitting from the zero-shot capabilities of SAM, we can combine the detected objects in Fig.~\ref{fig:heatmap_generation} (d) and the ground-truth ones in Fig.~\ref{fig:heatmap_generation} (e) to produce the final target heatmap in Fig.~\ref{fig:heatmap_generation} (f) to train the point decoder. Consequently, the target map can include as many objects as possible, and retain the ones missed by SAM at the same time.

Here, we compute the contour centers of all mask proposals predicted from uniform prompts, since we find that most point prompts may not be accurately located at the center of objects. It is worthwhile to note that there may be a few duplicated points for the same object; their mask proposals can be removed during post-processing.

We train the point decoder following~\cite{cheng2020higherhrnet}.
We splat all estimated keypoints into a heatmap $\mat{H}\in [0, 1]^{\frac{H}{s}\times \frac{W}{s}}$ using a Gaussian kernel $\mat{H}_{xy}=\exp(\frac{(sx-p_x)^2+(sy-p_y)^2}{2 \sigma^ 2})$, where $p\in\mathbb{R}^{2}$ is the keypoint and $\sigma=2$ is the standard deviation according to~\cite{cheng2020higherhrnet}. $x$ and $y$ are the coordinates on the heatmap. If the Gaussians of different objects overlap, we take the element-wise maximum~\cite{zhou2019objects}.
The point-wise mean squared error is employed for training:
\begin{align}
    \mathcal{L}_{\mathrm{point}}=\|\widehat{\mat{H}}-\mat{H}\|_2^2.
    \label{eq:heatmap_loss}
\end{align}
It is worthwhile to note that the quantization errors caused by output stride are not taken into account. The goal of the point decoder is to provide good class-agnostic point prompts for SAM, instead of an accurate object localization. SAM can segment certain objects as long as they are pointed out.

During inference, a $3\times 3$ max-pooling is applied to the heatmap to extract the peak key points, and the top $K$ of them with the scores above the threshold are selected. As a result, we can detect all points from $256\times 256$ grid with only $K=1000$, which has the same computational costs as $32\times 32$ grid points. In practice, there are fewer points than $K$ as studied in our experiments.

\subsection{Generalized Object Classification}
Given all proposals produced in Sec.~\ref{sec:object_localization}, this section aims to provide scores with respect to the image/text queries. In object counting, the image queries are cropped from input images according to the example bounding boxes. We construct the classification weights $\mat{W}\in \mathbb{R}^{C\times D}$ from the fixed CLIP language embeddings of class names or image embeddings of example boxes. $C$ is the arbitrary number of queries, $D$ is the dimension of CLIP embeddings and novel queries can be appended to the end of $\mat{W}$. The region features $\vct{r}$ are extracted from the image features processed by ROI align~\cite{he2017mask} and a two-layer MLP.
The object detector is supervised by the annotations in the image:
\begin{align}
    \mathcal{L}_{\mathrm{cls}}=\mathrm{BCE}(\mat{W}\vct{r}, \vct{c}),
    \label{eq:cls_loss}
\end{align}
where $\mathrm{BCE}$ is the binary cross-entroy loss following~\cite{zhou2022detecting}, and $\vct{c}$ is the ground-truth labels. $\vct{c}$ can be all zeros if the proposals are not matched with any ground-truth boxes.

In practice, this design yields unsatisfying results in generalizing to novel classes, as zero-shot capability of CLIP can be compromised when simply applying classification loss to the object classifier. Existing solutions include knowledge distillation~\cite{gu2021open} and enlarged vocabulary with image-caption training data~\cite{zhou2022detecting}. They are limited to object counting, which needs more discriminative representations since most scenes of object counting are crowded with small objects.

\paragraph{Hierarchical knowledge distillation.} We instead propose to align the region features and CLIP image embeddings of the hierarchical mask proposals from SAM. Similar to Eq.~\ref{eq:cls_loss}, for the mask proposals obtained from the same point, we build the classification weights from the CLIP image embeddings of cropped image regions. The region features are discriminated with corresponding CLIP embeddings according to their overlapping.
In doing so, the image encoder can be distilled to the classifier which meanwhile becomes more discriminative. This loss can be written as:
\begin{align}
    \mathcal{L}_{\mathrm{kd}}=\frac{1}{M}\sum_{i=1}^M \mathrm{BCE}(\mat{W}'\vct{r}^{(i)}, \vct{c}'),
    \label{eq:kd_loss}
\end{align}
where $M$ is the number of proposals of each point, $\mat{W}'\in \mathbb{R}^{M\times D}$ is the CLIP embeddings of image regions, and $\vct{c}'\in \mathbb{R}^{M}$ is filled 1 if the IoU between two proposals is larger than 0.5, otherwise 0. It is found that SAM usually fails to segment small objects in crowded scenes. To this end, we opt to use an additional $16\times 16$ box around each point to improve the segmentation of small objects, and only the first mask is selected. 
We note that the image regions and corresponding CLIP embeddings can be prepared in advance before training, similar to~\cite{gu2021open}. 
The visual illustration is shown in supplementary Fig.~\ref{fig:cc}.

\subsection{Training and Inference}
\label{sec:training}

The training loss function of our framework is the combination of Eqs.~\eqref{eq:heatmap_loss}, \eqref{eq:cls_loss}, and  \eqref{eq:kd_loss}:
\begin{align}
    \mathcal{L}=\mathcal{L}_{\mathrm{point}} + \mathcal{L}_{\mathrm{cls}} + \mathcal{L}_{\mathrm{kd}}.
\end{align}
During inference, non-maximum suppression is applied to all proposals to remove duplicate proposals, and the object counts are the number of detected bounding boxes with a score larger than the predefined threshold.

\section{Experiments}
\label{sec:exp}
\begin{figure*}[h]
    \centering
    \includegraphics[width=1.0\linewidth]{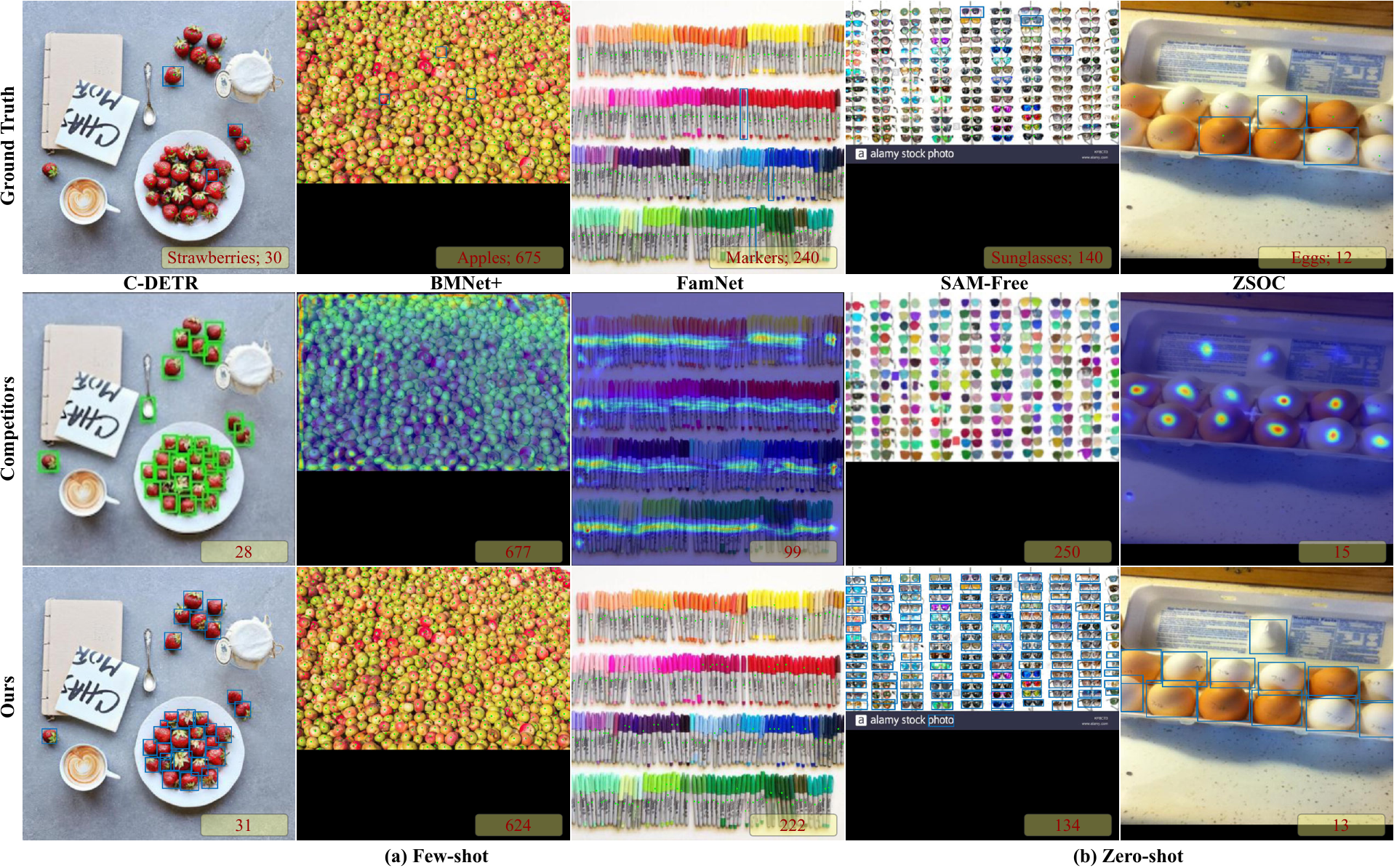}
    \caption{
        Qualitative comparisons for (a) few-shot~(the first 3 columns) and (b) zero-shot~(the last 2 columns) object counting.
        Only final points are placed in the second and third columns due to crowded predicted boxes.
        Zoom in for better view.
    }
    \vspace{-5pt}
    \label{fig:qualitative_comparisons}
\end{figure*}

\subsection{Implementation Details}

\paragraph{Datasets.}
For object counting, FSC-147 dataset~\cite{ranjan2021learning} is used to evaluate our method. It includes 6135 images of 147 categories, where the categories do not overlap between different splits. For object detection, FSC-147 and FSCD-LVIS~\cite{gupta2019lvis} annotated by~\cite{nguyen2022few} are employed. Since there are no box annotations in FSC-147 training data, we generate the pseudo-labels through the ground-truth point prompts. Although the pseudo-labels may be noisy, we find that it is sufficient to train a good classification network.

\paragraph{Training details.}
\methodname is trained for 50k iterations with a mini-batch size of 32, Adam optimizer~\cite{kingma2014adam} with a fixed learning rate of $10^{-4}$, weight decay of $10^{-5}$, $\beta_1=0.9$ and $\beta_2=0.99$. We utilize ViT-H~\cite{dosovitskiy2020image} of SAM and ViT-B of CLIP.
The point decoder is initialized by the mask decoder of SAM.
$K=1000$ and the threshold of heatmap is 0.05. 256 proposals and 16 pairs of each sample are randomly selected to train the classifier with $\mathcal{L}_{\mathrm{cls}}$ and $\mathcal{L}_{\mathrm{kd}}$, making sure 25\% positive proposals. No augmentation is used. The IoU threshold of NMS is $0.5$.

\paragraph{Evaluation metrics.}
For object counting, we adopt two standard metrics in the literature, including Mean Average Error (MAE) and Root Mean Squared Error (RMSE). In particular, $\mathrm{MAE}=\frac{1}{N} \sum_{i=1}^{N}\left|y_{i}-\hat{y}_{i}\right|$ and $\mathrm{RMSE}=\sqrt{\frac{1}{N} \sum_{i=1}^{N}\left(y_{i}-\hat{y}_{i}\right)^{2}}$, where $N$ is the number of samples, $y$ and $\hat{y}$ are the ground-truth and predicted object counts. We also reported Normalized Relative Error (NAE) and Squared Relative Error (SRE) in supplementary Tab.~\ref{tab:nae_sre_results}. For object detection, we use AP and AP50 strictly following~\cite{nguyen2022few}.

\subsection{Experimental Results}

\subsubsection{Qualitative Results} 

Fig.~\ref{fig:qualitative_results} showcases example results on few-shot/zero-shot object counting/detection.
Our \methodname has produced distinct detection results and accurate counts. \methodname can detect these small objects and discriminate well between the objects and background with the given example images/texts. However, our method is slightly not robust to the occlusion of objects. This is because SAM cannot distinguish occluded objects. Interestingly, \methodname can employ the text prompts to detect the objects accurately, even though given bad examples. For example, the example boxes of the deer in the last samples are only annotated around the head. We can address this problem by text prompts, whereas~\cite{xu2023zero} selects better example boxes. 
We show failure cases in supplementary Fig.~\ref{fig:qualitative_results_supp}.

In addition, Fig.~\ref{fig:qualitative_comparisons} presents the qualitative comparisons with few-shot object counting: C-DETR~\cite{nguyen2022few}, BMNet+~\cite{shi2022represent} and FamNet~\cite{ranjan2021learning}, and zero-shot object counting: SAM-Free~\cite{shi2023training} and ZSOC~\cite{xu2023zero}. We directly referred the results from their published papers to avoid any potential bias of self-implementation. \methodname has more interpretable detection results than density maps and performs competitively at extremely crowded scenes. \methodname also presents superior and discriminative detections than C-DETR and SAM-Free.

\subsubsection{Quantitative Results}

We evaluate the proposed \methodname on the crowded class-agnostic counting dataset FSC-147~\cite{yang2018learning}, under the setting of few-shot and zero-shot object counting. As a detection-based counting method, we also evaluate on the object detection datasets, FSC-147~\cite{yang2018learning} and FSCD-LVIS~\cite{nguyen2022few}.

\paragraph{Results on few-shot object counting.} 
In the few-shot counting scenario, each image provides three bounding box annotations of exemplar objects, which are used to count the target object in this image. Example images are cropped from input image and used to extract CLIP image embeddings as classification weights. Tab.~\ref{tab:fewshot_results} shows the quantitative comparisons with recent state-of-the-art methods, including detection-based and density-based methods.

Our \methodname achieves comparable MAE and RMSE with state-of-the-art density-based methods such as CounTR~\cite{liu2022countr} and BMNet+~\cite{shi2022represent}. \methodname also shows significant improvements over the detection-based methods. FR~\cite{fan2020few}, FSOD~\cite{fan2020few} and C-DETR~\cite{nguyen2022few} detect all proposals based on the state-of-the-art Faster R-CNN~\cite{ren2015faster} or DETR~\cite{carion2020end}. Their performance is limited since there are no sufficient training data to enable the detection models for better generalization ability.
SAM~\cite{kirillov2023segment,ma2023can} and SAM-Free~\cite{shi2023training} segment all objects and compute their similarities with examples to identify desired objects. However, they use only $32\times 32$ grid point prompts, leading to the failure in detecting small objects. Our \methodname can address this problem with class-agnostic object localization for more accurate point prompts, and generalize well to novel categories under the help of CLIP.

\paragraph{Results on zero-shot object counting.} 
Similar to the few-shot setting, we use the CLIP text embeddings as classification weights when only known class names. The results are shown in Tab.~\ref{table:zeroshot_results}. We have reproduced ViLD~\cite{gu2021open} on the proposed class-agnostic localization~(CAL) for better comparisons. 
% A similar phenomenon in few-shot counting is observed in zero-shot settings.
ViLD significantly outperforms  SAM~\cite{kirillov2023segment,shi2023training} and SAM-Free~\cite{shi2023training}, validating the effectiveness of class-agnostic object localization. Replacing ViLD with the proposed classifier, the performance is further improved. We find that there exists a great gap between few-shot and zero-shot settings, which may be caused by the ambiguous class names in FSC-147 dataset, such as the go pieces labeled as `go game'.

\paragraph{Results on object detection.} 
We evaluate the performance of object detection under both few-shot and zero-shot settings on the test set of FSC-147 and FSCD-LVIS~\cite{fan2020few}. The results are reported in Tab.~\ref{tab:detection_results}. Our \methodname achieves almost $2\times$ performance improvements compared to C-DETR~\cite{nguyen2022few}, due to the use of SAM and the proposed components. SAM is trained on large-scale datasets so that it can generalize well to extract accurate mask proposals for unseen categories.
CAL+ViLD would degenerate the performance, but is still better than baselines, demonstrating the effectiveness of class-agnostic object localization~(CAL).
% Although ViLD uses the proposed class-agnostic object localization, \methodname still outperforms it with the help of other components\shan{confused...why vild uses the proposed object localization}.
% We note that the evaluation on zero-shot object detection may be not as reliable as few-shot detection, as the class names are not matched well with the true bounding boxes.
% Interestingly, we find opposite behavior between FSC-147 and FSCD-LVIS datasets under zero-shot/few-shot settings; that is, the zero-shot performance on FSCD-LVIS is better than the few-shot one\shan{confused as well}. We think it is caused by the bad image or text prompts in different datasets.
Interestingly, \methodname behaves oppositely between two datasets under zero-shot/few-shot settings; that is, \methodname performs better with few shots than zero shots on FSC-147, versus on FSCD-LVIS. We think it happens since the example images in FSCD-LVIS are worse than text prompts.

\begin{table}[]
    \centering
    \scalebox{0.93}{
    \begin{tabular}{lrrrrr}
    \toprule
     & \multicolumn{2}{c}{Val set} & & \multicolumn{2}{c}{Test set} \\
     \cmidrule{2-3} \cmidrule{5-6}
     & MAE$_{\downarrow}$ & RMSE$_{\downarrow}$ & & MAE$_{\downarrow}$ & RMSE$_{\downarrow}$ \\
     \midrule
     \demph{GMN~\cite{lu2019class}} & \demph{29.66} & \demph{89.81} & & \demph{26.52} & \demph{124.57} \\
     \demph{MAML~\cite{finn2017model}} & \demph{25.54} & \demph{79.44} & & \demph{24.90} & \demph{112.68} \\
     \demph{FamNet~\cite{ranjan2021learning}} & \demph{23.75} & \demph{69.07} & & \demph{22.08} & \demph{99.54}  \\
     \demph{BMNet+~\cite{shi2022represent}} & \demph{15.74} & \demph{58.53} & & \demph{14.62} & \demph{91.83}  \\
     \demph{CounTR~\cite{liu2022countr}} & \demph{13.13} & \demph{49.83} & & \demph{11.95} & \demph{91.23}  \\
     \midrule
     FR~\cite{fan2020few} & 45.45 & 112.53 & & 41.64 & 141.04 \\
     FSOD~\cite{fan2020few} & 36.36 & 115.00 & & 32.53 & 140.65 \\
     C-DETR~\cite{nguyen2022few} & - & - & & 16.79 & 123.56 \\
     SAM~\cite{kirillov2023segment,ma2023can} & 31.20 & 100.83 & & 27.97 & 131.24 \\
     SAM-Free~\cite{shi2023training} & - & - & & 19.95 & 132.16 \\
     \textbf{Ours} & \textbf{15.31} & \textbf{68.34} & & \textbf{13.05} & \textbf{112.86} \\
    \bottomrule
    \end{tabular}
    }
    \vspace{-5pt}
    \caption{Results on few-shot object counting. The first and second parts contain the density-based and detection-based methods. We note that detecting objects for counting is much more challenging than predicting density maps. The best results are shown in bold.}
    \label{tab:fewshot_results}
\end{table}

\begin{table}[]
    \centering
    \scalebox{0.92}{
    \begin{tabular}{lrrrrr}
    \toprule
     & \multicolumn{2}{c}{Val set} & & \multicolumn{2}{c}{Test set} \\
     \cmidrule{2-3} \cmidrule{5-6}
     & MAE$_{\downarrow}$ & RMSE$_{\downarrow}$ & & MAE$_{\downarrow}$ & RMSE$_{\downarrow}$ \\
     \midrule
     \demph{RepRPN-C~\cite{ranjan2022exemplar}} & \demph{30.40} & \demph{98.73} & & \demph{28.32} & \demph{128.76}  \\
     \demph{CounTR~\cite{liu2022countr}} & \demph{17.40} & \demph{70.33} & & \demph{14.12} & \demph{108.01}  \\
     \demph{ZSOC~\cite{xu2023zero}} & \demph{26.93} & \demph{88.63} & & \demph{22.09} & \demph{115.17}  \\
     \midrule
     SAM~\cite{kirillov2023segment,shi2023training} & - & - & & 42.48 & 137.50 \\
     SAM-Free~\cite{shi2023training} & - & - & & 24.79 & 137.15 \\
     CAL+ViLD~\cite{gu2021open} & 28.81 & 111.13 & & 17.80 & 130.88 \\
     \textbf{Ours} & \textbf{23.90} & \textbf{100.33} & & \textbf{16.58} & \textbf{129.77} \\
    \bottomrule
    \end{tabular}
    }
    \vspace{-5pt}
    \caption{Results on zero-shot object counting.}
    \label{table:zeroshot_results}
\end{table}

\begin{table}[]
    \centering
    \scalebox{0.93}{
    \begin{tabular}{lrrrrr}
    \toprule
     & \multicolumn{2}{c}{FSC-147} & & \multicolumn{2}{c}{FSCD-LVIS} \\
     \cmidrule{2-3} \cmidrule{5-6}
     & AP$_{\uparrow}$ & AP50$_{\uparrow}$ & & AP$_{\uparrow}$ & AP50$_{\uparrow}$ \\
     \midrule
     C-DETR~\cite{nguyen2022few} & 22.66 & 50.57 & & 4.92 & 14.49 \\
     \textbf{Ours} & \textbf{43.53} & \textbf{74.64} & & \textbf{22.37} & \textbf{42.56} \\
     \midrule
     CAL+ViLD~\cite{gu2021open} & 40.56 & 67.21 & & 19.67 & 39.33 \\
     \textbf{Ours} & \textbf{41.14} & \textbf{69.03} & & \textbf{23.93} & \textbf{44.54} \\
    \bottomrule
    \end{tabular}
    }
    \vspace{-5pt}
    % \caption{Results on few-shot~(first part) and zero-shot~(second part) object detection.}
    \caption{Results on few-shot/zero-shot object detection.}
    \label{tab:detection_results}
\end{table}

\subsection{Ablation Study}
We evaluate different components in terms of few-shot detection/counting on FSC-147, and report the results of test set in Tab.~\ref{tab:ablation_study}. The baseline is detailed in Sec.~\ref{sec:problem_formulation}.

\begin{enumerate}[label=\color{red!70!black}(\roman*),wide,labelindent=0pt,itemsep=0ex,parsep=0pt,topsep=0pt]
    \item\label{bl_1} \textbf{Ablation on the localization.}\quad We compare our proposed class-agnostic object localization with two variants: (\textbf{i}) grid points and (\textbf{ii}) point decoder trained with only ground-truth points. The same trained classification network is used for fair comparisons. The object localization from the point decoder can improve both detection and counting performance compared to grid points. However, the point decoder could overfit the training categories if only using ground-truth annotations, which thus harms its generalization performance to unseen categories in testing data. \methodname combines all possible objects to alleviate this problem. 
    \item\label{bl_2} \textbf{Ablation on the SAM.}\quad We use ViT-H as the default SAM in this paper. By comparing different versions of SAM, as expected, heavier SAM introduces more improvements. However, our method of ViT-B version still performs better than C-DETR~\cite{nguyen2022few} as in Tab~\ref{tab:fewshot_results}.
    \item\label{bl_3} \textbf{Ablation on $\mathcal{L}_{\mathrm{kd}}$.}\quad We remove $\mathcal{L}_{\mathrm{kd}}$ from \methodname and find that the performance drops. $\mathcal{L}_{\mathrm{kd}}$ can distill the knowledge of CLIP to the object classification network, and thus greatly improve the detection performance on unseen classes. It can further enable the object classification network to discriminate the hierarchical and small mask proposals from SAM. As a result, the counting performance, especially RMSE, is significantly improved.
    \item\label{bl_4} \textbf{Ablation on the computation costs.}\quad Compared to vanilla SAM in~\cite{shi2023training} that employs $32\times 32$ grid point prompts, \methodname only selects an average of 378/388 candidate points for each image in the FSC-147 test/val sets. These points are selected from $256\times 256$ heatmaps, 64 times more than SAM. Detailed discussion is in supplementary Sec.~\ref{sec:costs}.
    
    % \item\label{bl_4} \textbf{Ablation on the computation costs.}\quad It is acknowledged that our method is slower than traditional two-stage object detection or object counting methods if using the same backbone. The computation costs of our method mainly lie in the frequent inference of the mask decoder of SAM. However, compared to vanilla SAM in~\cite{shi2023training} that employs $32\times 32$ grid point prompts, the proposed class-agnostic localization significantly reduces the computation costs. Specifically, there are only an average of 378 and 388 candidate points for each image in the FSC-147 test and val sets. It is worthwhile to note that these points are selected from $256\times 256$ heatmaps, 64 times than $32\times 32$ grid points. In addition, the point decoder shares the same architecture as the mask decoder of SAM and only needs 1 inference for each image.
\end{enumerate}

\begin{table}[]
    \centering
    \scalebox{0.93}{
    \begin{tabular}{llrrrrr}
    \toprule
     & & \multicolumn{2}{c}{Detection} & & \multicolumn{2}{c}{Counting} \\
     \cmidrule{3-4} \cmidrule{6-7}
     & & AP$_{\uparrow}$ & AP50$_{\uparrow}$ & & MAE$_{\downarrow}$ & RMSE$_{\downarrow}$ \\
     \midrule
                & Baseline & 39.63 & 67.50 & & 21.24 & 129.62 \\
     \midrule
     \ref{bl_1} & Grid & 41.66 & 71.53 & & 17.15 & 121.17 \\
                & \begin{tabular}[l]{@{}l@{}}Heatmap\\(only GT)\end{tabular}  & 43.26 & 73.62 & & 16.24 & 123.96 \\
     \midrule
     \ref{bl_2} & ViT-B & 39.83 & 70.72 & & 16.41 & 120.40 \\
                & ViT-L & 42.58 & 73.18 & & 14.65 & 118.64 \\
     \midrule
     \ref{bl_3} & w/o $\mathcal{L}_{\mathrm{kd}}$ & 41.69 & 70.59 & & 14.57 & 127.95 \\
                % & +Boxes & 42.18 & 71.40 & & 13.53 & 127.39 \\
     \midrule
              & Ours & \textbf{43.53} & \textbf{74.64} & & \textbf{13.05} & \textbf{112.86} \\
    \bottomrule
    \end{tabular}
    }
    \caption{Ablation study of different components in \methodname.}
    \vspace{-5pt}
    \label{tab:ablation_study}
\end{table}

\subsection{Results on Large-scale Datasets}
Although most methods  evaluate their object counting performance on FSC-147 dataset due to a large number of objects in each image,  we further evaluate \methodname on two more practical, complex but sparse datasets: COCO~\cite{lin2014microsoft} and LVIS~\cite{gupta2019lvis} under open-vocabulary object detection. Specifically, the categories in testing data may be unseen by the detection model, and only class names are known during testing. We strictly follow the experimental settings in the state-of-the-art object detection method  Detic~\cite{zhou2022detecting}. On COCO, we report the AP50$_\mathrm{n}$ for novel classes and AP50 for all classes. Similarly, we report mask AP on LVIS, \ie, AP$^\mathrm{m}_{\mathrm{n}}$ and AP$^\mathrm{m}$ on the novel and all classes.
For fair comparisons, additional trained backbone in~\cite{zhou2022detecting} and caption data is used to train the classification network. 
% We also use the segmentations from SAM to compute the mask AP.
%As they are sparse, 
Note that we do not report the counting performance due to their nature of sparseness.

The results in Tab.~\ref{tab:large_ov_results} show that our \methodname achieves significant performance improvements over Detic.
\begin{table}[h]
    \centering
    \scalebox{0.90}{
    \begin{tabular}{lrrrrr}
    \toprule
     & \multicolumn{2}{c}{COCO} & & \multicolumn{2}{c}{LVIS} \\
     \cmidrule{2-3} \cmidrule{5-6}
     & AP50$_{\mathrm{n}}$$_{\uparrow}$ & AP50$_{\uparrow}$ & & AP$^\mathrm{m}_{\mathrm{n}}$$_{\uparrow}$ & AP$^\mathrm{m}$$_{\uparrow}$ \\
     \midrule
     \demph{OVR-CNN~\cite{zareian2021open}} & \demph{22.8} & \demph{46.0} & & \demph{-} & \demph{-} \\
     \demph{ViLD~\cite{gu2021open}} & \demph{27.6} & \demph{51.3} & & \demph{16.1} & \demph{22.5} \\
     \midrule
     Detic~\cite{zhou2022detecting} & 27.8 & 45.0 & & 17.8 & 26.8 \\
    %  \textbf{Ours} & \textbf{23.90} & \textbf{100.33} & & \textbf{16.58} & \textbf{129.77} \\
     \textbf{Ours} & \textbf{32.9} & \textbf{47.5} & & \textbf{20.6} & \textbf{30.8} \\
    \bottomrule
    \end{tabular}
    }
    \vspace{-5pt}
    \caption{
        Results on large-scale but sparse detection datasets.
        We strictly follow the settings in Detic~\cite{zhou2022detecting} for fair comparisons. The first part may use unfair conditions, \eg, training data; all results are adopted from their corresponding papers.
    }
    \label{tab:large_ov_results}
\end{table}
\section{Conclusion}

In this paper, we introduce \methodname, a generalized framework for few-shot/zero-shot object detection/counting.
\methodname follows the spirits: point, segment, and count, which synergizes the advantages of SAM and CLIP. It employs a class-agnostic object localization to provide good point prompts for SAM.
Extensive experiments validate the effectiveness of \methodname on both object detection/counting, including FSC-147, large-scale COCO, and LVIS datasets.
In the future, we will explore how to achieve fine-grained object counting inspired by current great success in multi-modal LLM.

\small{\noindent\textbf{Acknowledgement}\quad This work was supported in part by STI2030-Major Projects (No.~2021ZD0200204), National Natural Science Foundation of China (Nos.~62176059 and 62101136),  Shanghai Municipal Science and Technology Major Project (No.~2018SHZDZX01) and ZJLab, and  Shanghai Center for Brain Science and Brain-inspired Technology.}

\clearpage
%%%%%%%%% REFERENCES
{\small
\bibliographystyle{ieee_fullname}
\bibliography{ref}
}

% \clearpage
\clearpage
\setcounter{page}{1}
\maketitlesupplementary

\begin{figure}[h]
    \centering
    \includegraphics[width=0.5\linewidth]{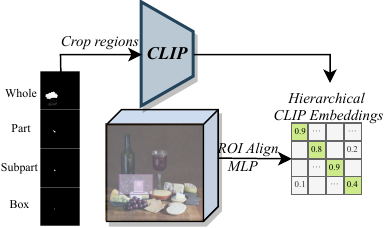}
    \caption{
    Illustration of the proposed hierarchical knowledge distillation. The hierarchical mask proposals from SAM of the same points are cropped from the input image, which are then fed into CLIP. The CLIP embeddings are used as classification weights to discriminate the classifier, which encourages discriminative classification among hierarchical proposals.
    }
    \label{fig:cc}
\end{figure}

\begin{figure*}[h]
  \centering
  \includegraphics[width=1.0\linewidth]{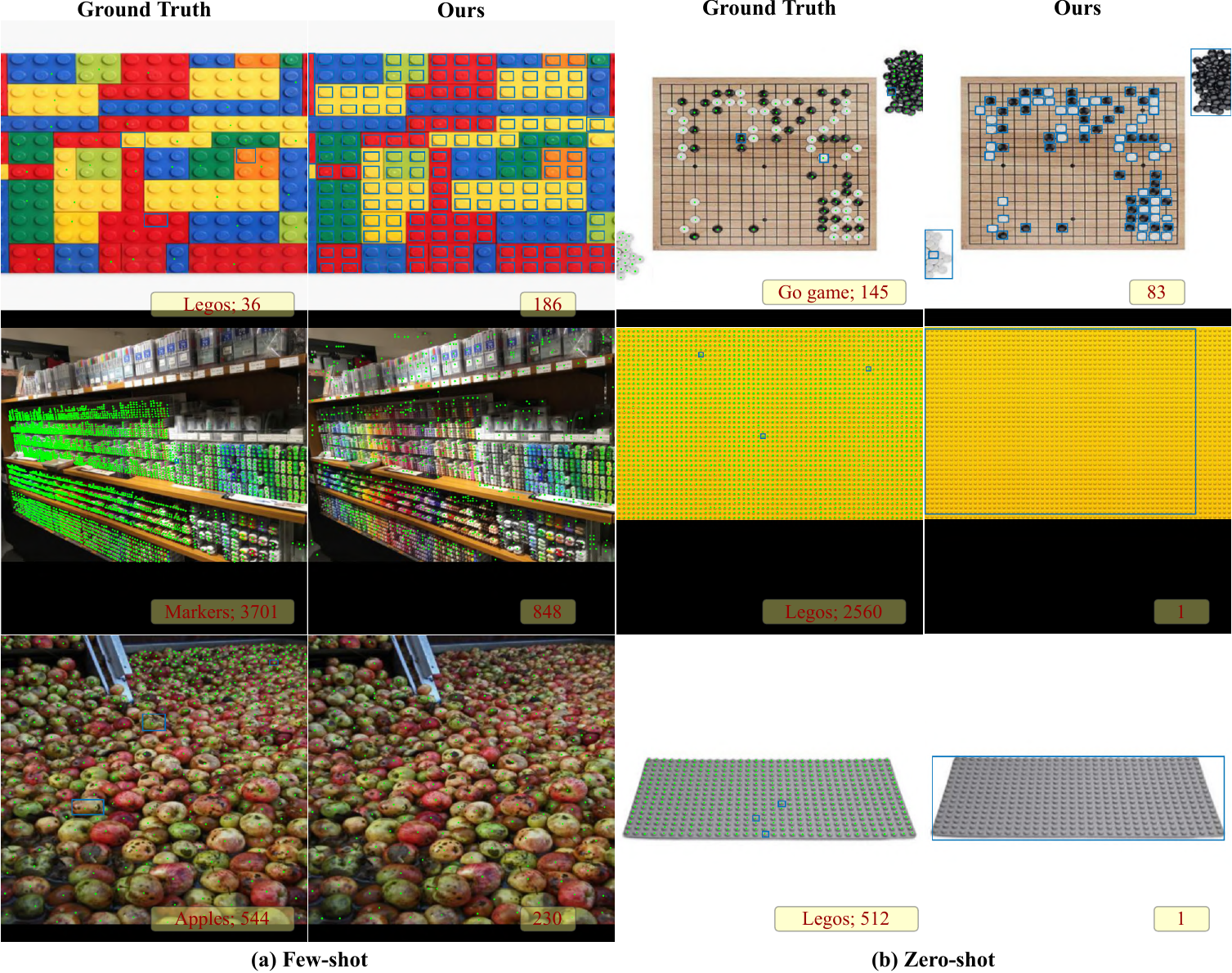}
  \caption{
      Failure cases for (a) few-shot and (b) zero-shot object counting and detection. Under the few-shot setting, \methodname may fail to identify target objects given wrong example images, and cannot detect objects on extremely crowded scenes. A similar phenomenon has been observed under the zero-shot setting. The imprecise text prompts, \eg, `go game' and `legos' cannot be used to distinguish the objects.
  }
  \label{fig:qualitative_results_supp}
\end{figure*}

This supplementary material provides the following extra content:
\begin{enumerate}
    \item Visual illustration of proposed hierarchical knowledge distillation in Fig.~\ref{fig:cc};
    \item Results of object counting in terms of NAE and SRE in Tab.~\ref{tab:nae_sre_results};
    \item Failure cases of proposed \methodname in Fig.~\ref{fig:qualitative_results_supp}.
    \item Ablation on the computation costs in Sec.~\ref{sec:costs}.
\end{enumerate}

\begin{table}[h]
  \centering
  \begin{tabular}{lrrrrr}
  \toprule
   & \multicolumn{2}{c}{Val set} & & \multicolumn{2}{c}{Test set} \\
   \cmidrule{2-3} \cmidrule{5-6}
   & NAE$_{\downarrow}$ & SRE$_{\downarrow}$ & & NAE$_{\downarrow}$ & SRE$_{\downarrow}$ \\
   \midrule
   SAM-Free~\cite{shi2023training} & - & - & & 0.29 & 3.80 \\
   C-DETR~\cite{nguyen2022few} & - & - & & 0.19 & 5.23 \\
   \textbf{Ours} & \textbf{0.22} & \textbf{2.99} & & \textbf{0.19} & \textbf{3.12} \\
   \midrule
    ZSOC~\cite{xu2023zero} & 0.36 & \textbf{4.26} & & 0.34 & 3.74  \\
   SAM-Free~\cite{shi2023training} & - & - & & 0.37 & 4.52 \\
   \textbf{Ours} & \textbf{0.32} & 4.31 & & \textbf{0.25} & \textbf{3.46} \\
  \bottomrule
  \end{tabular}
  \caption{
    Results on the few-shot~(first part) and zero-shot~(second part) object counting. We reported the results on FSC-147 in terms of Normalized Relative Error (NAE) and Squared Relative Error (SRE). $\mathrm{NAE}=\frac{1}{N} \sum_{j=1}^{N} \frac{\left|\hat{y}_{i}-y_{i}\right|}{y_{i}}$ and $\mathrm{SRE}=\sqrt{\frac{1}{N} \sum_{i=1}^{N} \frac{\left(\hat{y}_{i}-y_{i}\right)^{2}}{y_{i}}}$, where $y_{i}$ and $\hat{y}_{i}$ are GT and predicted counts. ZSOC belongs to density-based methods.
    Our proposed method achieves state-of-the-art methods on the two metrics.
    }
  \label{tab:nae_sre_results}
\end{table}

\section{Ablation on the computation costs.}
\label{sec:costs}
It is acknowledged that our method is slower than traditional two-stage object detection or object counting methods if using the same backbone. The computation costs of our method mainly lie in the frequent inference of the mask decoder of SAM. However, compared to vanilla SAM in~\cite{shi2023training} that employs $32\times 32$ grid point prompts, the proposed class-agnostic localization significantly reduces the computation costs. Specifically, there are only an average of 378 and 388 candidate points for each image in the FSC-147 test and val sets. It is worthwhile to note that these points are selected from $256\times 256$ heatmaps, 64 times than $32\times 32$ grid points. In addition, the point decoder shares the same architecture as the mask decoder of SAM and only needs 1 inference for each image.

\end{document}